\documentclass[letterpaper]{article} 
\usepackage{aaai2027}  
\nocopyright
\usepackage[hyphens]{url}  
\usepackage{graphicx} 
\usepackage{natbib}  
\usepackage{caption} 
\usepackage{algorithm}

\usepackage{newfloat}
\usepackage{listings}
\DeclareCaptionStyle{ruled}{labelfont=normalfont,labelsep=colon,strut=off} 
\floatstyle{ruled}
\newfloat{listing}{tb}{lst}{}
\floatname{listing}{Listing}

\usepackage{booktabs}

\usepackage{multirow}
\usepackage[table]{xcolor}
\usepackage{placeins}
\usepackage{algpseudocode}
\usepackage{booktabs}
\usepackage{tabularx}
\usepackage{arydshln}
\usepackage{subcaption}
\usepackage{caption}
\usepackage{graphicx}
\usepackage{amsmath}
\usepackage{amssymb}

\title{Xemo-Talker: Unlock Emotions Explicitly for \\ Audio-Driven Talking Portrait Synthesis}
\author {
    Chaolong Yang \textsuperscript{\rm 1,\rm 2,\rm 3}\equalcontrib,
    Yinuo Guo \textsuperscript{\rm 4}\equalcontrib,
    Kai Yao\textsuperscript{\rm 5},
    Yuyao Yan\textsuperscript{\rm 2},
    Jie Sun\textsuperscript{\rm 2}\corresponding,
    Guangliang Cheng\textsuperscript{\rm 1}, \\
    Shibin Wu\textsuperscript{\rm 6},
    Bin Dong\textsuperscript{\rm 7},
    Kaizhu Huang\textsuperscript{\rm 3}\corresponding
}

\affiliations {
    \textsuperscript{\rm 1}University of Liverpool,
    \textsuperscript{\rm 2}Xi’an Jiaotong-Liverpool University,
    \textsuperscript{\rm 3}Duke Kunshan University,
    \textsuperscript{\rm 4}Carnegie Mellon University,
    \textsuperscript{\rm 5}Ant Group,
    \textsuperscript{\rm 6}Pingan Technology,
    \textsuperscript{\rm 7}Ricoh Software Research Center\\
    Chaolong.Yang@liverpool.ac.uk, yinuog@andrew.cmu.edu, jiumo.yk@antgroup.com, yuyao.yan@xjtlu.edu.cn, Jie.Sun@xjtlu.edu.cn, Guangliang.Cheng@liverpool.ac.uk, wushibin37@163.com, bin.dong@dukekunshan.edu.cn, Kaizhu.Huang@dukekunshan.edu.cn
}

\begin{document}

\maketitle

\begin{abstract}
Precise emotion control in audio-driven talking heads remains a challenge due to the reliance on implicit emotion regulation in existing systems, which often leads to indirect and insufficient control. Additionally, training with explicit emotion-related losses across the entire motion space poses significant difficulties due to the inherent trade-off between accurate lip synchronization and fine-grained emotion control. In this paper, we reveal a key finding: although emotional cues are distributed throughout the motion space, concentrating discriminative supervision on less-principal components achieves a better emotion–lip synchronization balance, as principal components mainly encode high-energy articulation and pose variations. Building on this insight, we propose Xemo-Talker, which first learns a neutral speech-to-motion mapping for stable articulation and lip synchronization, and then introduces a lightweight emotion branch guided by less-principal subspace supervision. To enhance emotion control, we design a Tri-Loss consisting of inter-class separation, intra-class compactness, and less-principal contrastive learning. Given an audio input, a reference image, and an emotion label, Xemo-Talker achieves state-of-the-art emotion classification accuracy while maintaining competitive lip synchronization and high inference efficiency, with performance approaching that measured on real videos. The source code will be publicly available at \url{https://github.com/chaolongy/Xemo-Talker}.
\end{abstract}

\section{Introduction}
\label{sec:intro}

Audio-driven talking head synthesis has achieved substantial progress in lip synchronization, motion realism, and visual quality~\cite{prajwal2020lip,zhou2020makelttalk,zhou2021pose,zhang2023sadtalker,cui2024hallo2,cui2025hallo3,wang2025fantasytalking,fei2025skyreels,yang2025unlock,yang2025kdtalker++}, yet reliable emotion control remains challenging. Existing methods often learn affect implicitly within audio-to-motion networks~\cite{ji2022eamm,gururani2023space,xia2023gmtalker,zhang2025magictalk}, resulting in weak or inconsistent expressions. Recent approaches introduce explicit emotion-related objectives~\cite{tan2024edtalk,wang2025pc,tan2025disentangle}, but applying strong supervision uniformly across the motion space may interfere with articulation-related motion, creating a trade-off between emotional expressiveness and lip synchronization.

\begin{figure}[t]
\centering
\begin{subfigure}[t]{0.98\linewidth}
\centering
\includegraphics[width=\linewidth]{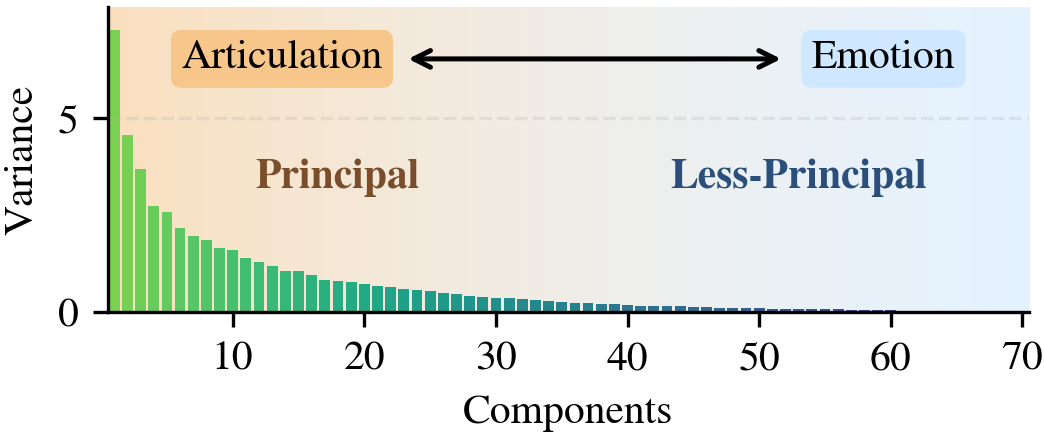}
\caption{Ordered PCA variance spectrum of the 70-D motion space.}
\label{fig:pca_variance}
\end{subfigure}\hfill
\begin{subfigure}[t]{0.98\linewidth}
\centering
\includegraphics[width=\linewidth]{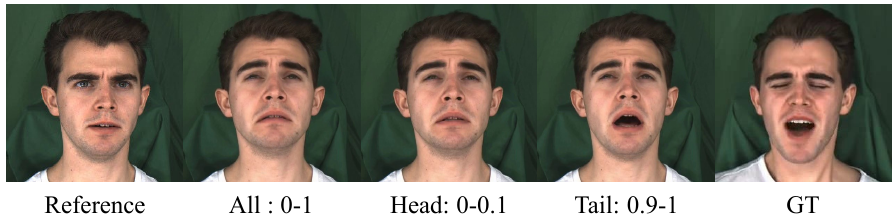}
\caption{Effect of emotion supervision on different PCA ranges.}
\label{fig:pca_qualitative}
\end{subfigure}
\caption{PCA analysis of the 70-D motion space. Principal components mainly capture high-energy articulation and pose, while the less-principal tail provides a lower-interference region for emotion supervision. Head and Tail denote normalized PCA ranges $0$--$0.1$ and $0.9$--$1$, respectively. PCA is computed on the MEAD training split.}
\label{fig:pca_combined}
\vspace{-10pt}
\end{figure}

Psychological studies show that facial expressions are composed of localized facial action units and that subtle temporal dynamics are important for emotion perception~\cite{ekman1978facial,ambadar2005deciphering}. Motivated by these findings, we investigate whether strong emotion supervision should be applied uniformly across the motion space or concentrated in directions that interfere less with dominant speech motion. As shown in Fig.~\ref{fig:pca_variance}, we apply PCA to per-frame 70-D motion parameters extracted by the frozen LivePortrait model~\cite{guo2024liveportrait} from the MEAD training set~\cite{wang2020mead}. The variance spectrum is highly skewed: principal components are dominated by high-energy articulation and head-pose variations, including jaw opening and large rigid movements, whereas tail components describe lower-amplitude facial changes. 
Although articulation and emotional cues are both distributed throughout the motion space, the low-variance tail provides a lower-interference region for strengthening emotion discrimination without directly perturbing dominant speech-related motion.

We verify this observation by applying emotion supervision to the full PCA space, the principal region, and the less-principal tail, as illustrated in Fig.~\ref{fig:pca_qualitative}. Full-space or principal-region supervision more strongly alters articulation-related mouth motion, whereas tail-focused supervision mainly refines subtle facial variations. Analytical experiments show that less-principal supervision achieves a better emotion--lip synchronization balance. We therefore do not claim that emotion exists only in the PCA tail; rather, concentrating discriminative supervision in this subspace reduces its interference with articulation- and pose-dominant directions.

Building on this insight, we propose Xemo-Talker, a two-stage framework for explicit emotion control. A Geometry Motion Predictor (GMP) first learns a neutral speech-to-motion mapping for stable articulation and lip synchronization. A Geometry Emotion Branch (GEB) then injects emotion-conditioned residual features through zero-initialized adapters while continuing to predict the complete 70-D motion. To strengthen emotion control, we design a subspace-aware Tri-Loss containing an inter-class classification term, an intra-class prototype-alignment term, and a less-principal contrastive term. The first two organize the global emotion space, while the last concentrates strong discriminative supervision on low-variance directions to reduce interference with dominant speech motion.

In summary, Xemo-Talker addresses the trade-off between emotion expressiveness and articulation fidelity by separating neutral speech-motion learning from emotion refinement and concentrating discriminative supervision in a lower-interference PCA subspace. Our contributions are:
\begin{itemize}
\item We empirically analyze emotion supervision across different PCA regions and find that less-principal supervision provides a better emotion--lip synchronization trade-off than full-space or principal-region supervision.

\item We propose Xemo-Talker, a two-stage framework that separates neutral speech-motion learning from explicit emotion refinement through a Geometry Motion Predictor and a Geometry Emotion Branch.

\item We design a subspace-aware Tri-Loss combining inter-class classification, intra-class prototype alignment, and less-principal contrastive learning to improve emotion discrimination with reduced interference to dominant speech motion.

\item Extensive experiments show state-of-the-art emotion accuracy, strong identity preservation, competitive lip synchronization, and efficient inference.
\end{itemize}

\newcommand{\cmark}{\ensuremath{\checkmark}}
\newcommand{\xmark}{\ensuremath{\times}}

\section{Related Work}
\label{sec:rw}

\subsection{Audio-Driven Talking Head Generation}
Audio-driven talking head generation has progressed from direct image synthesis to explicit motion modeling and diffusion-based generation. Wav2Lip~\cite{prajwal2020lip} synthesizes lip-synchronized frames, while MakeItTalk~\cite{zhou2020makelttalk}, PC-AVS~\cite{zhou2021pose}, and SadTalker~\cite{zhang2023sadtalker} predict facial landmarks, disentangled motion, or 3D coefficients before rendering. Geometry-domain approaches such as KDTalker~\cite{yang2025unlock} separate identity-agnostic motion prediction from portrait rendering, improving efficiency and controllability.

Recent diffusion-based methods, including Hallo2~\cite{cui2024hallo2}, Hallo3~\cite{cui2025hallo3}, FantasyTalking~\cite{wang2025fantasytalking}, SkyReels-Audio~\cite{fei2025skyreels}, and Memo~\cite{zheng2024memo}, substantially improve visual realism, temporal coherence, and motion diversity. However, these methods mainly target general expressiveness rather than independent categorical emotion control or the interaction between emotion modulation and speech articulation.

\subsection{Emotional Talking Head Generation}

\begin{table}[t]
\centering
\scriptsize
\setlength{\tabcolsep}{2.8pt}
\renewcommand{\arraystretch}{1.10}
\resizebox{\columnwidth}{!}{
\begin{tabular}{cccccc}
\toprule
\multirow{2}{*}{Category} &
\multirow{2}{*}{Method} &
Emotion &
Explicit emotion &
Factor &
Less-principal \\
&
&
guidance &
supervision &
decoupling &
supervision \\
\midrule

Foundation &
Memo &
None &
\xmark &
\xmark &
\xmark \\

Foundation &
Hallo3 &
None &
\xmark &
\xmark &
\xmark \\

\midrule

Implicit &
GMTalker &
Emotion label &
\xmark &
\xmark &
\xmark \\

\midrule

Explicit &
EAT &
Label/Text &
\cmark &
\cmark &
\xmark \\

Explicit &
EDTalk &
Audio/Video &
\cmark &
\cmark &
\xmark \\

Explicit &
DICE-Talk &
Audio/Video &
\cmark &
\cmark &
\xmark \\

Explicit &
EmoCAST &
Text &
\cmark &
\cmark &
\xmark \\

\midrule
\rowcolor{gray!20}
Explicit &
Xemo-Talker &
Emotion label &
\cmark &
\cmark &
\cmark \\

\bottomrule
\end{tabular}}
\caption{Positioning of representative talking-head methods.}
\label{tab:rw_position}
\vspace{-10pt}
\end{table}

Methods without dedicated emotion-discrimination objectives typically introduce affect through reference videos, emotion labels, or latent codes. EAMM~\cite{ji2022eamm} and SPACE~\cite{gururani2023space} transfer emotional motion or style from references, while GMTalker~\cite{xia2023gmtalker} and MagicTalk~\cite{zhang2025magictalk} model emotion through latent distributions or audio-expression correlations. These methods provide flexible emotion conditioning, but affect is learned jointly with speech motion, often resulting in entangled and less consistent control.

Explicit approaches improve controllability through emotion-specific modules, structured objectives, or factorized representations. EAT~\cite{gan2023eat} uses lightweight emotional adaptation, while EmoHead~\cite{shen2025emohead} and EmoCAST~\cite{jiang2025emocast} introduce semantic parameterization or text guidance. EDTalk~\cite{tan2024edtalk}, PC-Talk~\cite{wang2025pc}, DICE-Talk~\cite{tan2025disentangle}, EmotiveTalk~\cite{wang2025emotivetalk}, and Cafe-Talk~\cite{chen2025cafe} further separate emotion from articulation, identity, pose, or other facial controls. Although these designs strengthen emotional expressiveness, they do not explicitly consider how emotion supervision should be distributed across motion directions with different variance and articulation relevance.

As summarized in Tab.~\ref{tab:rw_position}, Xemo-Talker addresses the issue from a subspace-aware perspective. It first learns emotion-agnostic speech motion for articulation and then introduces emotion-conditioned refinement. Global classification and prototype alignment organize emotion representations, while contrastive discrimination is applied to the less-principal PCA projection, improving emotion control with reduced interference to articulation-dominant motion.

\section{Method}
\label{sec:method}

\begin{figure*}[t]
  \centering
  \includegraphics[width=\linewidth]{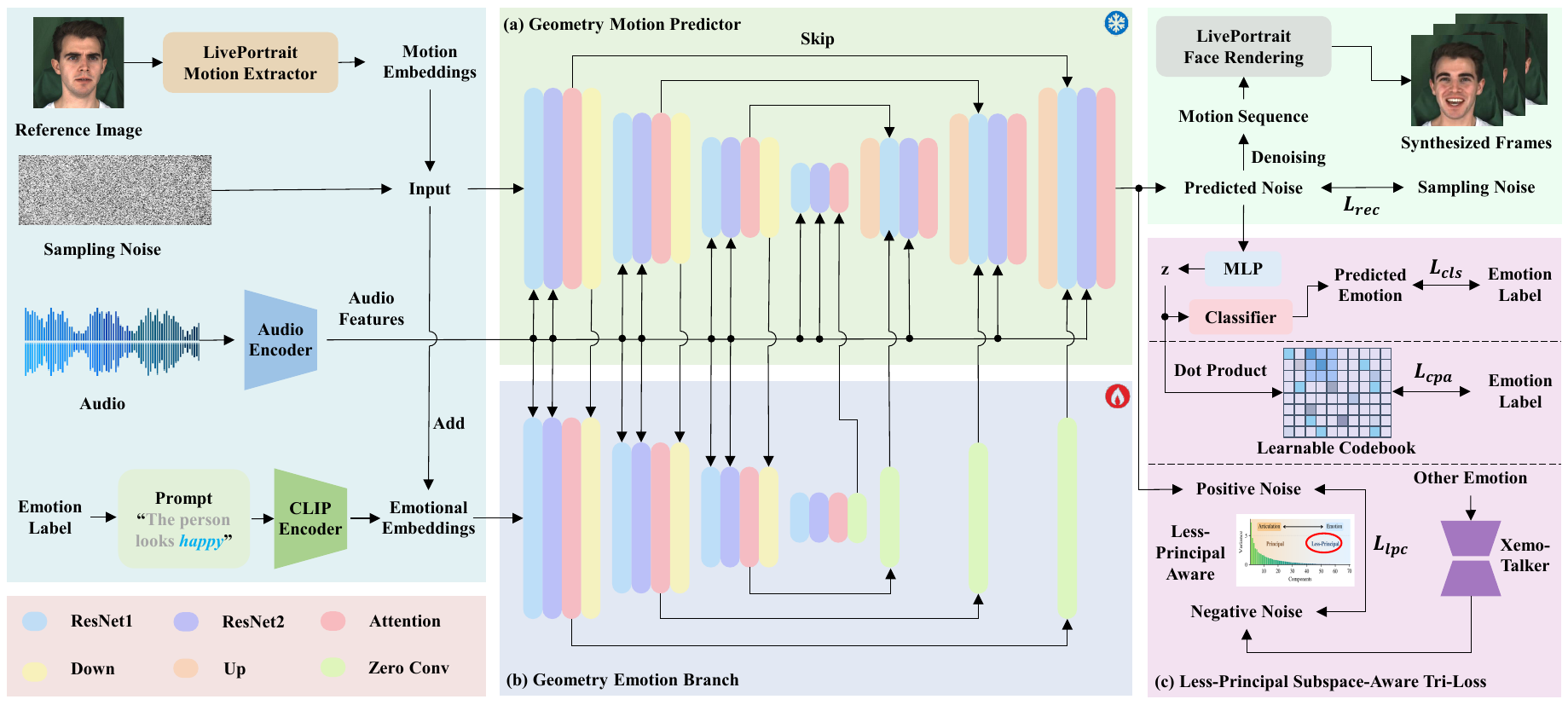}
  \caption{
  Overview of Xemo-Talker.
  (a) The Geometry Motion Predictor learns emotion-agnostic audio-driven motion
  through diffusion reconstruction, producing the complete 70-D facial motion
  sequence.
  (b) With the predictor frozen, the Geometry Emotion Branch encodes the target
  emotion and injects multi-scale residual features through zero-initialized
  adapters to refine the motion prediction.
  (c) The subspace-aware Tri-Loss organizes global emotion representations using
  classification and prototype alignment, while applying contrastive
  discrimination to the less-principal PCA projection.
  The predicted motion is converted into deformed keypoints and rendered by the
  frozen LivePortrait renderer~\cite{guo2024liveportrait}.}
  \label{fig:overview}

\end{figure*}

\subsection{Overview and Motion Representation}
\label{sec:overview}

Given a reference image $I_r$, an audio sequence $A_{1:T}$, and an emotion label $y$, Xemo-Talker generates an emotion-controllable talking portrait. As shown in Fig.~\ref{fig:overview}, it consists of a Geometry Motion Predictor (GMP), a Geometry Emotion Branch (GEB), and a less-principal subspace-aware Tri-Loss. Frozen LivePortrait~\cite{guo2024liveportrait} provides the motion representation and renders the predicted motion into video frames. For frame $i$, the motion vector contains deformation $\delta_i\in\mathbb{R}^{63}$, scale $s_i\in\mathbb{R}$, translation $t_i\in\mathbb{R}^{3}$, and rotation $r_i\in\mathbb{R}^{3}$:
\begin{equation}
m_i=
\left[
s_i,\,
t_i^\top,\,
r_i^\top\,
\delta_i^\top,
\right]^\top
\in\mathbb{R}^{70}.
\label{eq:motion_representation}
\end{equation}
The motion sequence is
$X_0=[m_1,\ldots,m_T]^\top\in\mathbb{R}^{T\times70}$.
Given canonical keypoints $x_c$, the deformed keypoints are:
\begin{equation}
x_{d,i}
=
s_i\bigl(x_cR(r_i)+\delta_i\bigr)+t_i,
\label{eq:keypoint_deformation}
\end{equation}
where $R(\cdot)$ converts rotation parameters into a rotation matrix. Training proceeds in two stages. GMP first learns an emotion-agnostic audio-to-motion mapping using reconstruction supervision. GMP is then frozen, and the GEB injects multi-scale emotion-conditioned residuals to refine the motion prediction. Global classification and prototype alignment organize emotion representations, while less-principal contrastive learning strengthens emotion discrimination with reduced interference to articulation-dominant motion.

\subsection{Geometry Motion Predictor}
\label{sec:GMP}

The Geometry Motion Predictor (GMP) learns an emotion-agnostic audio-to-motion mapping without explicit emotion supervision. Following KDTalker~\cite{yang2025unlock}, we construct a reference condition $C_r$ from the canonical keypoints and source motion extracted by the frozen LivePortrait~\cite{guo2024liveportrait} Motion Extractor. GMP is implemented as a temporal U-Net that processes the noised motion sequence, diffusion timestep, and reference condition. Audio features $A_{1:T}$ are extracted by a Wav2Lip-based encoder~\cite{prajwal2020lip}, injected through Feature-wise Linear Modulation (FiLM), and fused with motion features using cross-modal attention. The network predicts noise for the complete 70-D motion sequence.

Given a clean motion sequence $X_0$, Gaussian noise $\epsilon\sim\mathcal{N}(0,I)$, and timestep $\tau$, the noised motion is $X_\tau=\sqrt{\bar{\alpha}_\tau}X_0+ \sqrt{1-\bar{\alpha}_\tau}\epsilon$. Let $c=(A_{1:T},C_r)$ denote the audio and reference conditions. GMP is optimized by the noise-prediction objective
\begin{equation}
\mathcal{L}_{\mathrm{rec}}
=
\mathbb{E}_{X_0,\epsilon,\tau}
\left[
\left\|
\epsilon-U_\theta(X_\tau,\tau;c)
\right\|_2^2
\right],
\label{eq:l_rec}
\end{equation}
where $U_\theta$ denotes the GMP noise predictor. At inference, DDIM sampling~\cite{song2021ddim} recovers the complete motion sequence from Gaussian noise, which is then converted into deformed keypoints and rendered by the frozen LivePortrait renderer.

\subsection{Geometry Emotion Branch}
\label{sec:GEB}

In the second stage, the pretrained GMP is frozen and augmented with a Geometry Emotion Branch (GEB) for explicit emotion control. GEB mirrors the temporal U-Net structure of GMP and receives the same noised motion, reference condition, and audio features. The target emotion label is inserted into a predefined prompt and encoded by the CLIP text encoder~\cite{radford2021learning} as a global emotion condition. Audio and emotion conditions are injected independently, with audio features modulating the residual blocks through Feature-wise Linear Modulation (FiLM).

To preserve the speech-motion prior learned by GMP, we introduce Cross-Stage Feature Injection (CSFI). At each encoder scale $l$, the frozen GMP feature $g_l$ is added to the corresponding GEB feature $e_l$. A zero-initialized $1\times1$ convolution $Z_l$ then produces an emotion-conditioned residual, which modulates the corresponding GMP decoder feature $d_l$:
\begin{equation}
\tilde e_l=e_l+g_l,
\qquad
\tilde d_l=d_l+Z_l(\tilde e_l).
\label{eq:geb_injection}
\end{equation}
The residual injection is applied at the bottleneck. Zero initialization preserves the original GMP behavior at the beginning of training and enables emotion refinement. The modulated GMP decoder predicts the complete 70-D motion sequence rather than a tail-only motion residual. GEB is optimized with the reconstruction objective and the subspace-aware Tri-Loss to strengthen emotion control while reducing interference with articulation-dominant motion.

\subsection{Less-Principal Subspace-Aware Tri-Loss}
\label{sec:Tri-Loss}

To enhance emotion control with less interference to articulation-dominant motion, we introduce a Tri-Loss comprising emotion classification, prototype alignment, and less-principal contrastive learning. Given noisy motion $X_{\tau}$ at timestep $\tau$ and predicted noise $\hat{\epsilon}_{\theta}^{y}$ conditioned on emotion label $y$, we recover the clean-motion estimate using the standard diffusion inversion. Let $\hat{\delta}_{i}^{y}\in\mathbb{R}^{63}$ denote its deformation component at frame $i$. We temporally aggregate the deformation sequence and project it into a 128-D normalized emotion representation:
\begin{equation}
z
=
\operatorname{Norm}
\left[
f_{\mathrm{proj}}
\left(
\frac{1}{T}
\sum_{i=1}^{T}
\hat{\delta}_{i}^{y}
\right)
\right]
\in\mathbb{R}^{128},
\label{eq:emotion_embedding}
\end{equation}
where $T$ is the sequence length, $f_{\mathrm{proj}}$ is a learnable projection network, and $\operatorname{Norm}$ denotes $\ell_2$ normalization.

\noindent\textbf{Global Emotion Organization.}
Let $E$ be the number of emotion categories. We use a linear classifier with $W_{\mathrm{cls}}\in\mathbb{R}^{E\times128}$ and $b_{\mathrm{cls}}\in\mathbb{R}^{E}$, together with a learnable prototype codebook $P\in\mathbb{R}^{E\times128}$. Let $\widetilde{P}$ denote the row-normalized codebook. The two losses are
\begin{equation}
\mathcal{L}_{\mathrm{cls}}
=\operatorname{CE}(W_{\mathrm{cls}}z+b_{\mathrm{cls}},y),
\quad
\mathcal{L}_{\mathrm{cpa}}
=\operatorname{CE}(\tau_p^{-1}z\widetilde{P}^{\top},y),
\label{eq:global_emotion_losses}
\end{equation}
where $\operatorname{CE}$ denotes cross-entropy and $\tau_p$ is the prototype temperature.

\noindent\textbf{Less-Principal Contrastive Learning.}
We compute PCA on the 70-D training motions. Let $\mu\in\mathbb{R}^{70}$ be the PCA mean and $U_{\mathrm{tail}}\in\mathbb{R}^{7\times70}$ contain the lowest-variance $10\%$ of the PCA directions. We define
\begin{equation}
\Phi_{\mathrm{tail}}(x)
=
\operatorname{vec}
\left[
(x-\mu)U_{\mathrm{tail}}^{\top}
\right],
\label{eq:tail_projection}
\end{equation}
where $\operatorname{vec}(\cdot)$ flattens the temporal projection.

For each label $y$, we sample a different label $y^{-}\neq y$. The two forward passes share all inputs except the emotion condition. For sample $n$, we obtain
\begin{equation}
a_n=\Phi_{\mathrm{tail}}(\hat{\epsilon}_{\theta,n}^{y}),
\quad
b_n=\operatorname{sg}\!\left[
\Phi_{\mathrm{tail}}(\hat{\epsilon}_{\theta,n}^{y^{-}})
\right],
\label{eq:tail_pair}
\end{equation}
where $\operatorname{sg}(\cdot)$ denotes stop-gradient. For a batch of $B$ samples, the less-principal contrastive loss is
\begin{equation}
\mathcal{L}_{\mathrm{lpc}}
=
\frac{1}{B}
\sum_{n=1}^{B}
\frac{
a_n^{\top}b_n
}{
\left\lVert a_n\right\rVert_{2}
\left\lVert b_n\right\rVert_{2}
}.
\label{eq:l_lpc}
\end{equation}
Minimizing $\mathcal{L}_{\mathrm{lpc}}$ reduces the similarity between different emotion conditions in the less-principal subspace.

\noindent\textbf{Overall Objective.}
The second-stage objective is
\begin{equation}
\mathcal{L}
=
\mathcal{L}_{\mathrm{rec}}
+
\lambda_{\mathrm{cls}}\mathcal{L}_{\mathrm{cls}}
+
\lambda_{\mathrm{cpa}}\mathcal{L}_{\mathrm{cpa}}
+
\lambda_{\mathrm{lpc}}\mathcal{L}_{\mathrm{lpc}}.
\label{eq:l_total}
\end{equation}
The $\lambda$ terms balance the emotion objectives. During the second stage, GMP remains frozen, while GEB and the auxiliary emotion modules are jointly optimized. The auxiliary modules are removed during inference.

\section{Experiments}
\label{sec:exp}

\subsection{Experimental Setup}
\label{sec:exp_setup}

\noindent\textbf{Datasets.}
The first-stage Geometry Motion Predictor is trained on emotion-unlabeled talking-head data from VoxCeleb~\cite{Nagrani19} and HDTF~\cite{zhang2021flow}, containing 9,594 and 1,963 clips, respectively. The second-stage Geometry Emotion Branch is trained on MEAD~\cite{wang2020mead}, which contains 60 speakers performing 30 sentences under eight emotion categories and three intensity levels. Following prior work~\cite{tan2024edtalk}, 43 speakers are used for training, while M003, M030, W009, and W015 are reserved for testing. All PCA statistics and emotion evaluators are obtained using the MEAD training split.

\noindent\textbf{Evaluation Metrics.}
We evaluate lip synchronization, visual quality, identity preservation, and emotion accuracy. Lip synchronization is measured by LSE-C and LSE-D from SyncNet~\cite{prajwal2020lip}, where higher LSE-C and lower LSE-D indicate better audio--visual alignment. Visual quality is evaluated using Fr\'echet Inception Distance (FID)~\cite{heusel2017gans} and Cumulative Probability of Blur Detection (CPBD)~\cite{narvekar2011no}, where lower FID and higher CPBD are preferred. Identity preservation is measured by cosine similarity (CSIM) between ArcFace~\cite{deng2019arcface} embeddings of the reference image and generated frames. Following EAT~\cite{gan2023eat}, we fine-tune Emotion-FAN~\cite{meng2019frame} on the MEAD training split and report top-1 emotion accuracy, denoted as $\mathrm{Acc}_{\mathrm{emo}}$.

\begin{table*}[t]
\setlength{\aboverulesep}{0pt}
\setlength{\belowrulesep}{0pt}
\centering
\resizebox{0.9\textwidth}{!}{%
\begin{tabular}{l|cc|ccc|c|c}
\toprule[1.2pt]
\multirow{2}{*}{\centering Method}  
& \multicolumn{2}{c|}{Lip Synchronization} 
& \multicolumn{4}{c|}{Video Quality} 
& Emotion Accuracy \\ 
\cline{2-8} 
& LSE-C $\uparrow$ & LSE-D $\downarrow$ 
& FID $\downarrow$ & CPBD $\uparrow$ & CSIM $\uparrow$ & Avg. Rank $\downarrow$
& $\text{Acc}_{\textit{emo}}$ $\uparrow$ \\ 
\hline

Real Video    & 8.04 & 7.52 & 0.00 & 0.47 & 1.00 & -- & 85.38 \\ \hline

EAT~\cite{gan2023eat} [ICCV'23] 
& \textbf{7.29} & 8.35 & \underline{45.79} & 0.20 & \textbf{0.68} & \textbf{2.00} & 75.43 \\

EDTalk~\cite{tan2024edtalk} [ECCV'24]
& 6.75 & \underline{8.30} & 73.00 & 0.16 & 0.62 & 4.33 & 81.02 \\

DICE-Talk~\cite{tan2025disentangle} [MM’25]
& 6.61 & 8.76 & 74.87 & \textbf{0.31} & \underline{0.63} & \underline{3.00} & 42.03 \\

EmoCAST~\cite{jiang2025emocast} [arXiv'25]
& \underline{7.08} & 8.37 & \textbf{43.35} & 0.20 & 0.59 & \underline{3.00} & \underline{84.26} \\

\rowcolor{gray!20}
\textbf{Xemo-Talker (Ours)}
& 6.37 & \textbf{8.21} & 69.67 & \underline{0.30} & \textbf{0.68} & \textbf{2.00} & \textbf{85.28} \\

\textcolor{gray}{\quad - Stage-I-only}
& \textcolor{gray}{6.15} & \textcolor{gray}{8.02} 
& \textcolor{gray}{65.48} & \textcolor{gray}{0.28} & \textcolor{gray}{0.48} & \textcolor{gray}{--}
& \textcolor{gray}{15.84} \\

\bottomrule[1.2pt]
\end{tabular}%
}
\caption{Quantitative comparison with the state-of-the-art methods on MEAD~\cite{wang2020mead} test set. Best in \textbf{bold} and second best in \underline{underline}. Avg. Rank denotes the average ranking over FID, CPBD, and CSIM.}
\label{tab:quantitative}
\vspace{-10pt}
\end{table*}

\begin{figure*}[t]
  \centering
  \includegraphics[width=0.9\linewidth]
  {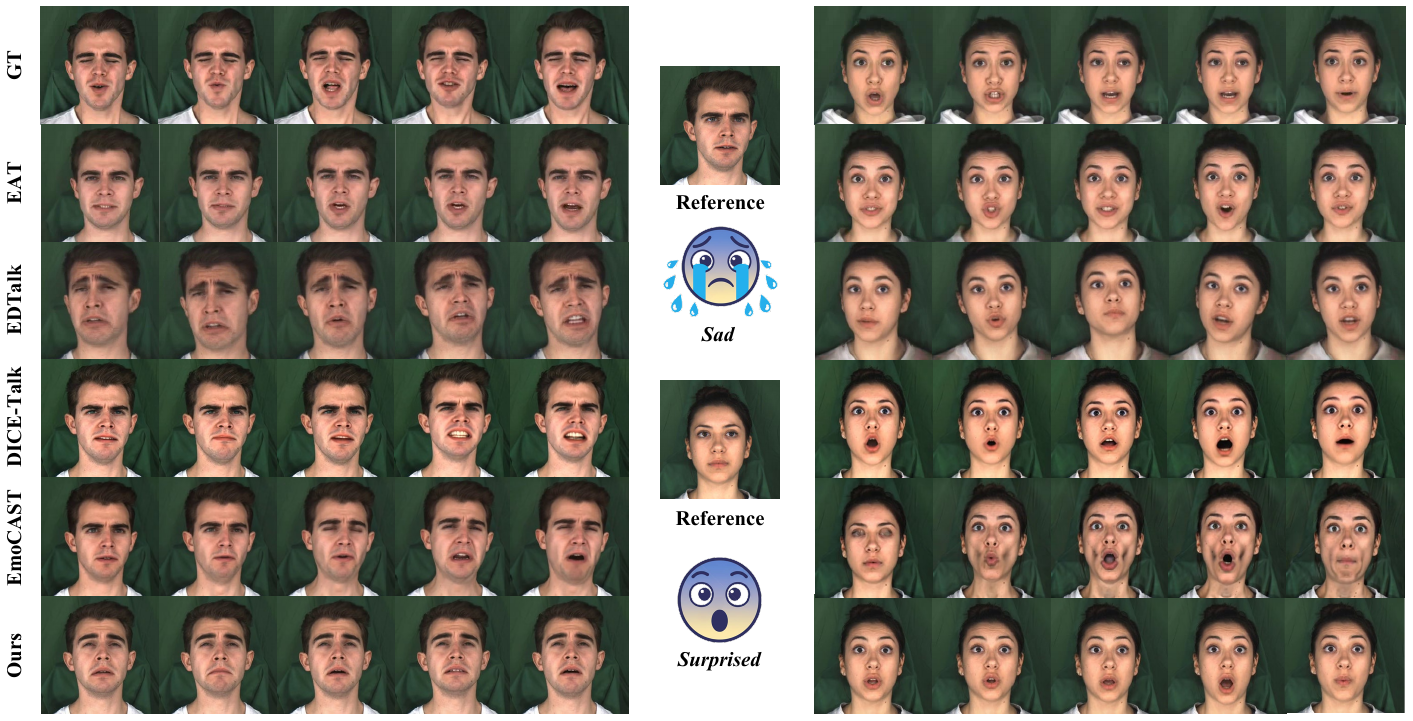}
  \caption{Qualitative comparison on the MEAD test set for sad and surprised emotions. Xemo-Talker produces clear emotional variations while preserving facial details and source identity.}
  \label{fig:qualitative_comparison_sad_sur}
  \vspace{-10pt}
\end{figure*}

\noindent\textbf{Implementation Details.}
Xemo-Talker contains a 75.4M-parameter GMP and a 40.3M-parameter GEB. LivePortrait is frozen throughout training, and GMP is frozen in Stage II. Each sample contains $T=64$ frames represented by 70-D motion vectors and processed at $256\times256$ resolution, with audio features extracted by a pretrained Wav2Lip-based encoder. We use 1,000 diffusion timesteps for training and 50-step DDIM~\cite{song2021ddim} sampling for evaluation. Both stages are trained for 500,000 iterations with a batch size of 64 using AdamW~\cite{kingma2014adam}, an initial learning rate of $2\times10^{-6}$, linear warmup, and cosine decay. The PCA basis is computed from 70-D motion parameters extracted only from the MEAD training split. The final $10\%$ of the ordered PCA spectrum, corresponding to $K=7$ directions, is used for less-principal supervision. We set $\lambda_{\mathrm{cls}}=\lambda_{\mathrm{cpa}}=1$ and $\lambda_{\mathrm{lpc}}=0.1$. All experiments use a single NVIDIA RTX~4090 GPU.

\subsection{Comparison with State-of-the-Art Methods}
\label{sec:sota_comparison}

\noindent\textbf{Quantitative Comparison.}
Table~\ref{tab:quantitative} compares Xemo-Talker with emotional talking-head methods on MEAD. Xemo-Talker achieves the highest emotion accuracy among methods at $85.28\%$, outperforming EmoCAST by $1.02\%$ percentage points and approaching the real-video reference of $85.38\%$. It obtains the best LSE-D score of $8.21$, although its LSE-C remains lower than several baselines. These results indicate a favorable emotion-lip synchronization trade-off rather than consistent superiority across all synchronization metrics.

For visual and identity quality, Xemo-Talker achieves a CPBD of $0.30$, close to the best result of $0.31$, and ties for the highest CSIM at $0.68$. Consequently, it obtains the best tied average rank of \textbf{$2.00$}. Its FID of $69.67$ is higher than those of EAT and EmoCAST, indicating weaker dataset-specific appearance matching. Nevertheless, the CPBD and CSIM results show that Xemo-Talker preserves image sharpness and source identity while improving emotion control.

The Stage-I-only result isolates the effect of emotion training. Although GMP is frozen in Stage II, GEB refines complete 70-D motion prediction rather than masking lip-related dimensions. Stage-II training increases LSE-C from $6.15$ to $6.37$, while LSE-D slightly worsens from $8.02$ to $8.21$. In contrast, $\mathrm{Acc}_{\mathrm{emo}}$ increases from $15.84\%$ to $85.28\%$, a gain of $69.44$ percentage points, while CSIM improves from $0.48$ to $0.68$. These results show that Stage II  strengthens emotion control with only a small change in lip synchronization.

\noindent\textbf{Qualitative Comparison.}
Fig.~\ref{fig:qualitative_comparison_sad_sur} compares generated results for sad and surprised emotions. In the shown examples, Xemo-Talker produces clear emotion-specific facial variations while maintaining image sharpness and identity-related details. EAT~\cite{gan2023eat} and EDTalk~\cite{tan2024edtalk} exhibit relatively blurred facial details in some frames, while DICE-Talk~\cite{tan2025disentangle} produces weaker sad expressions that appear closer to neutral. EDTalk and EmoCAST~\cite{jiang2025emocast} also show noticeable changes in facial appearance for some identities. Overall, Xemo-Talker achieves a favorable balance among emotional expressiveness, visual clarity, and identity preservation in these examples.

\subsection{User Studies}
A user study was conducted to evaluate the performance of all methods. We select two people from the test set for each emotion and compare the results from all five models. In total, there are sixteen videos, each paired with three questions asking users to choose the best one for emotion accuracy, lip synchronization, and overall quality. Twenty people participated in this evaluation, and the results are shown in Table~\ref{tab:questionnaire_results}. Our model achieves the highest scores in emotion and overall quality, and the second-highest score in lip synchronization. This aligns with our experiments showing that Xemo-Talker achieves strong emotional expressiveness while maintaining competitive lip synchronisation, yielding a balanced result. DICE-Talk~\cite{tan2025disentangle} scores high in lip synchronization, but its lip movements are overly exaggerated, leading to lower overall quality. More user study details can be seen in the supplemental material.

\begin{table}[t]
\centering
\centering
\setlength{\aboverulesep}{0pt}
\setlength{\belowrulesep}{0pt}
\resizebox{0.8\linewidth}{!}{
\begin{tabular}{l|c|c|c}
\toprule
Model & Emotion & Lip & Overall \\
\hline
EAT~\cite{gan2023eat} & 20.31 & 15.00 & 21.56 \\
EDTalk~\cite{tan2024edtalk} & 11.56 & 8.75 & 10.62 \\
DICE-Talk~\cite{tan2025disentangle} & 13.75 & \textbf{32.81} & 16.56 \\
EmoCAST~\cite{jiang2025emocast} & \underline{25.00} & 17.50 & \underline{24.06} \\
\rowcolor{gray!20}
\textbf{Xemo-Talker (Ours)} & \textbf{29.38} & \underline{25.94} & \textbf{27.19} \\
\bottomrule
\end{tabular}
}
\caption{User study results (\%) across three evaluation metrics. Best in \textbf{bold} and second best in \underline{underline}.}
\label{tab:questionnaire_results}
\end{table}

\begin{table}[t]
\centering
\resizebox{0.8\linewidth}{!}{
\begin{tabular}{l|cc|c}
\toprule
\multirow{2}{*}{Variant}
& \multicolumn{2}{c|}{Lip Synchronization}
& Emotion Accuracy \\
\cline{2-4}
& LSE-C $\uparrow$
& LSE-D $\downarrow$
& $\text{Acc}_{\textit{emo}}$ $\uparrow$ \\
\hline

w/o $\mathcal{L}_{\mathrm{cls}}$
& \textbf{6.91}
& \textbf{8.09}
& 72.89 \\

w/o $\mathcal{L}_{\mathrm{cpa}}$
& 6.27
& 8.73
& 78.88 \\

w/o $\mathcal{L}_{\mathrm{lpc}}$
& 6.23
& 8.78
& 79.09 \\

\hline

w/o CSFI
& 6.36
& 8.21
& 82.23 \\

\hline

\rowcolor{gray!20}
Full model
& 6.37
& 8.21
& \textbf{85.28} \\

\bottomrule
\end{tabular}}
\caption{Ablation study of the Tri-Loss terms and CSFI. Each variant removes one component from the full model.}
\label{tab:as_main}
\vspace{-6pt}
\end{table}

\subsection{Analytical Experiments}
\label{subsec:anaExp}

\noindent\textbf{Ablation of Core Components.}
Table~\ref{tab:as_main} evaluates the three Tri-Loss terms and Cross-Stage Feature Injection (CSFI) through leave-one-out ablations. Removing $\mathcal{L}_{\mathrm{cls}}$ causes the largest decrease in emotion accuracy, from $85.28\%$ to $72.89\%$, confirming its importance for inter-class separation. Removing $\mathcal{L}_{\mathrm{cpa}}$ or $\mathcal{L}_{\mathrm{lpc}}$ reduces emotion accuracy to $78.88\%$ and $79.09\%$, respectively, showing the complementary roles of prototype organization and less-principal discrimination. Without CSFI, emotion accuracy decreases by $3.05$ percentage points, while LSE-C and LSE-D remain nearly unchanged. Although removing $\mathcal{L}_{\mathrm{cls}}$ improves lip-synchronization metrics, it weakens emotion control, highlighting the trade-off between articulation and emotional expressiveness. The full model achieves the highest emotion accuracy with limited variation in lip synchronization.

\begin{figure}[t]
  \centering
  \includegraphics[width=0.9\linewidth]
  {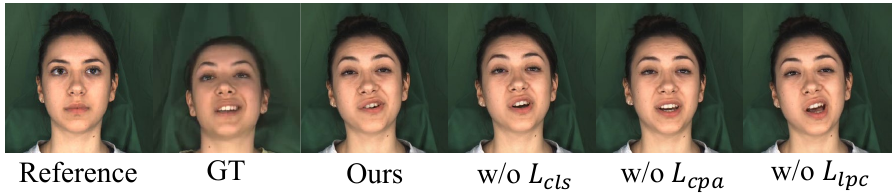}
  \caption{Qualitative results of the ablation study for Tri-loss. The emotion category is Contempt.}
  \label{fig:ablation_study_qualitative}
  \vspace{-8pt}
\end{figure}

The qualitative results in Fig.~\ref{fig:ablation_study_qualitative} illustrate the complementary effects of the three objectives. Without $\mathcal{L}_{\mathrm{cls}}$, the contempt expression becomes less distinctive, while removing $\mathcal{L}_{\mathrm{cpa}}$ weakens its intensity. Without $\mathcal{L}_{\mathrm{lpc}}$, the mouth region exhibits less coherent deformation. In comparison, the complete Tri-Loss produces a clearer emotion direction and more consistent local facial details in the shown example.

\begin{table}[t]
\centering
\resizebox{0.9\linewidth}{!}{
\begin{tabular}{l|cc|c}
\toprule
\multirow{2}{*}{Method}
& \multicolumn{2}{c|}{Lip Synchronization}
& Emotion Accuracy \\
\cline{2-4}
& LSE-C $\uparrow$
& LSE-D $\downarrow$
& $\text{Acc}_{\textit{emo}}$ $\uparrow$ \\
\hline

$Full_{(0-1)}$
& 6.28
& 8.34
& 75.84 \\

\hline

$Head_{(0-0.5)}$
& 6.36
& 8.32
& \textbf{79.49} \\

$Tail_{(0.5-1)}$
& \textbf{6.44}
& \textbf{8.20}
& 78.88 \\

\hline

$Head_{(0-0.3)}$
& 6.23
& 8.36
& 78.38 \\

$Tail_{(0.7-1)}$
& \textbf{6.39}
& \textbf{8.23}
& \textbf{79.09} \\

\hline

$Head_{(0-0.1)}$
& 6.31
& 8.32
& 79.80 \\

\rowcolor{gray!20}
$Tail_{(0.9-1)}$
& \textbf{6.51}
& \textbf{8.14}
& \textbf{80.51} \\

\bottomrule
\end{tabular}}
\caption{Effect of Less-Principal Supervision.}
\label{tab:pca_slice_pairs}
\vspace{-6pt}
\end{table}

\noindent\textbf{Effect of Less-Principal Supervision.}
To examine how emotion supervision affects lip motion, we apply $\mathcal{L}_{\mathrm{lpc}}$ to different PCA ranges while disabling $\mathcal{L}_{\mathrm{cls}}$ and $\mathcal{L}_{\mathrm{cpa}}$. Principal directions contain high-energy articulation and pose variations, whereas less-principal directions interfere less with speech-related lip motion. As shown in Table~\ref{tab:pca_slice_pairs}, tail supervision consistently achieves better LSE-C and LSE-D than the corresponding head ranges, while maintaining comparable or higher emotion accuracy in most settings. The final $10\%$ tail yields the best joint result, reaching $80.51\%$ emotion accuracy, 6.51 LSE-C, and 8.14 LSE-D. This confirms that concentrating explicit emotion supervision on less-principal directions strengthens emotion control while reducing interference with lip synchronization.

\begin{figure*}[t]
  \centering
  \includegraphics[width=0.85\linewidth]{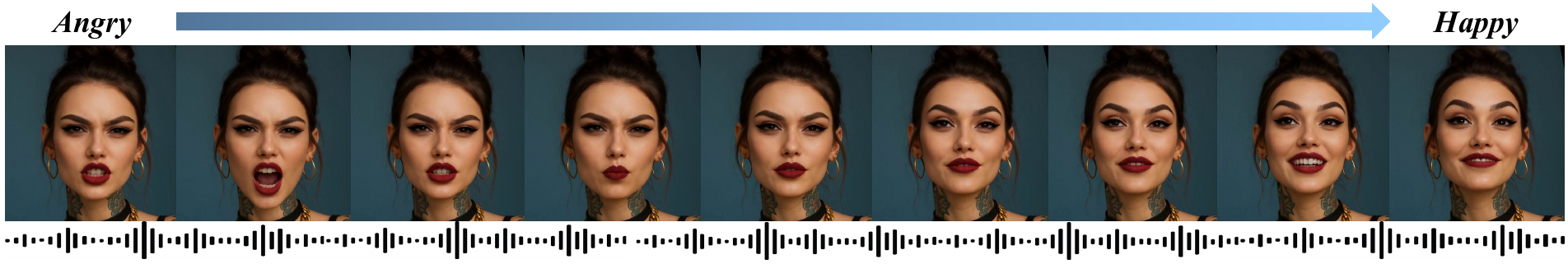}
  \caption{Cross emotion transition from angry to happy with fixed audio.}
  \label{fig:ablation_study_cross_emotion}
\end{figure*}

\begin{figure*}[t]
\centering
\includegraphics[width=0.85\textwidth]{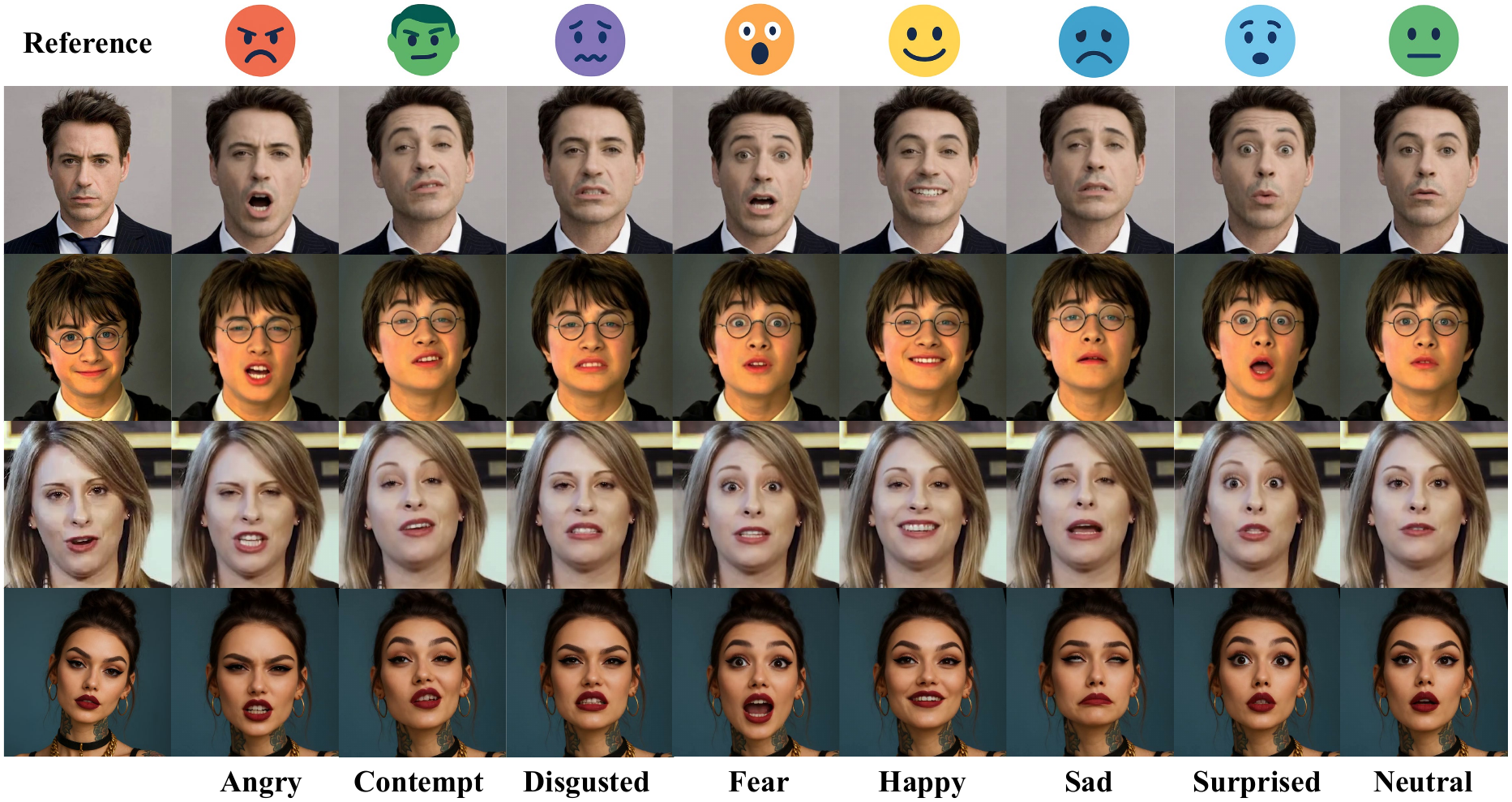}
\captionof{figure}{Xemo-Talker results across diverse emotions. The figure shows the model's ability to generate clear, expressive, and identity-consistent faces for all eight emotions in challenging out-of-domain scenarios.}
\label{fig:Xemo-Talker_ood}
\end{figure*}

\noindent\textbf{Inference Efficiency.}
Xemo-Talker uses a single DDIM sampling chain, where the frozen GMP and GEB jointly predict the complete 70-D motion sequence. The PCA projection and auxiliary emotion heads are used only during training. As shown in Table~\ref{tab:speed}, the 10-step setting reaches $27.02$ FPS, faster than all compared methods, while achieving $84.77\%$ emotion accuracy, slightly surpassing EmoCAST at $84.26\%$. The default 50-step setting achieves $14.32$ FPS and $85.28\%$ accuracy, remaining substantially faster than EDTalk, DICE-Talk, and EmoCAST while providing the highest emotion accuracy among the compared methods. These results demonstrate a favorable speed--accuracy trade-off.

\begin{table}[h]
\centering
\setlength{\aboverulesep}{0pt}
\setlength{\belowrulesep}{0pt}
\resizebox{0.8\linewidth}{!}{
\begin{tabular}{c|c|c}
\toprule
Method
& FPS $\uparrow$
& $\text{Acc}_{\textit{emo}}$ $\uparrow$ \\
\hline

EAT~\cite{gan2023eat}
& 22.06
& 75.43 \\

EDTalk~\cite{tan2024edtalk}
& 5.38
& 81.02 \\

DICE-Talk~\cite{tan2025disentangle}
& 0.73
& 42.03 \\

EmoCAST~\cite{jiang2025emocast}
& 0.41
& 84.26 \\

\hline

Ours (10 steps)
& \textbf{27.02}
& 84.77 \\

Ours (30 steps)
& 18.27
& 84.57 \\

Ours (50 steps)
& 14.32
& 85.28 \\

Ours (100 steps)
& 9.26
& \textbf{85.38} \\

\bottomrule
\end{tabular}}
\caption{Inference speed and $\text{Acc}_{\textit{emo}}$ on a single RTX~4090.}
\label{tab:speed}
\vspace{-6pt}
\end{table}

\begin{figure}[t]
  \centering
  \includegraphics[width=0.8\linewidth]{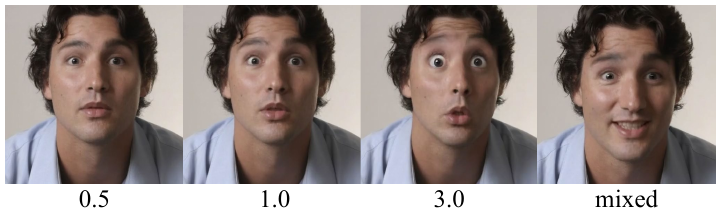}
  \caption{Emotion intensity scaling and interpolation between surprised and
  happy.}
  \label{fig:emo_control}
  \vspace{-6pt}
\end{figure}

\noindent \textbf{Fine-grained Emotion Control.}
Our method supports fine-grained emotion control beyond discrete classes. Intensity is adjusted by scaling the emotion residual with $\alpha\in\{0.5,1,3\}$, and the emotion condition can be specified per frame for timeline control. Mixed emotions are obtained by blending embeddings $e=(1-\beta)e_a+\beta e_b$, $\beta\in[0,1]$. As no standard metric exists for intensity/mixing controllability, we report qualitative results in Fig.~\ref{fig:emo_control}.

\noindent \textbf{Cross-Emotion Transition.}
We evaluate Xemo-Talker under time-varying emotion labels with fixed audio. The first half of the sequence is labeled angry and the second half happy. As shown in Fig.~\ref{fig:ablation_study_cross_emotion}, the model maintains a consistent angry expression with stable articulation and head motion, then transitions near the midpoint by relaxing the upper face. The second segment exhibits a clear happy expression with natural smiles synchronized to the audio. These results show that Xemo-Talker follows temporal emotion changes while preserving lip synchronization and identity stability.

\noindent \textbf{Out-of-Domain Generalization.}
Fig.~\ref{fig:Xemo-Talker_ood} evaluates Xemo-Talker on challenging out-of-domain cases beyond the MEAD test set. Despite diverse identities, camera setups, and recording conditions, the model produces clear, expressive faces across all emotions while maintaining identity consistency, plausible head pose and eye motion, and distinct emotional cues. This suggests that the geometry-emotion design and Tri-Loss supervision generalize beyond a single dataset to realistic deployment scenarios.

\section{Conclusion}
\label{sec:con}

We present Xemo-Talker, a geometry-domain framework for explicit and fine-grained emotion control in audio-driven talking portrait synthesis. Our analysis shows that emotional cues are distributed throughout the motion space, while emotion-dependent differences can be more effectively emphasized along less-principal directions with less interference to lip-related motion. Accordingly, Xemo-Talker first learns an emotion-agnostic speech-to-motion mapping and then refines the complete motion prediction using an emotion-conditioned branch. Its subspace-aware Tri-Loss organizes global emotion representations while selectively strengthening less-principal emotion discrimination. Experiments demonstrate state-of-the-art emotion accuracy with competitive lip synchronization, identity preservation, and inference efficiency, providing an effective balance between emotional expressiveness and articulation fidelity.

\bibliography{aaai2027}


\clearpage
\setcounter{page}{1}

\makeatletter
\@topnewpage[
  \begin{minipage}{\textwidth}
    \centering
    {\LARGE\bfseries
    Xemo-Talker: Unlock Emotions Explicitly for\\
    Audio-Driven Talking Portrait Synthesis\par}

    \vspace{0.8em}

    {\large\bfseries Supplementary Material\par}

    \vspace{1.2em}
  \end{minipage}
]
\makeatother

\appendix
\setcounter{figure}{7}
\renewcommand{\thesection}{\Alph{section}}

\section{Additional Qualitative Comparisons}
\label{sec:more_qc}

\begin{figure*}[t]
  \centering
  \includegraphics[width=1\linewidth]{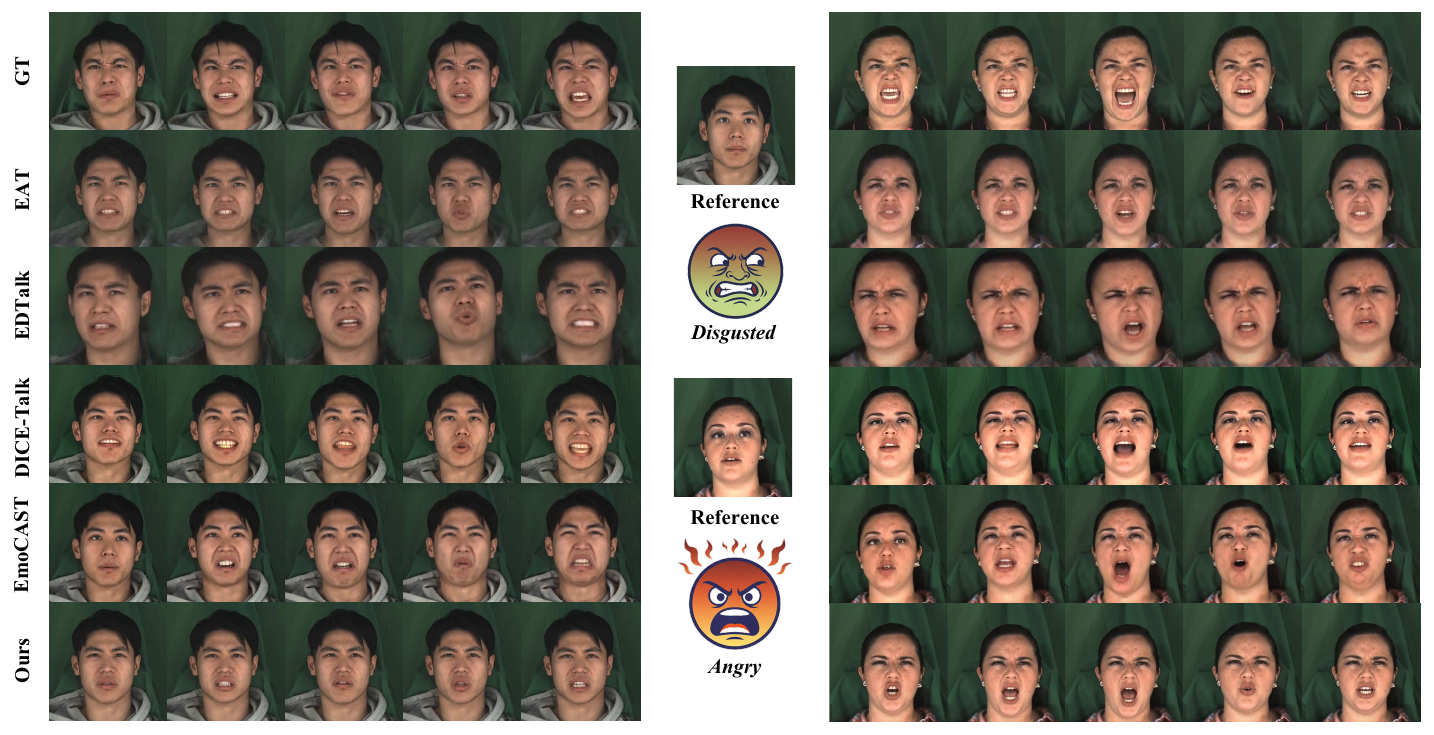}
  \caption{Qualitative comparison on the MEAD~\cite{wang2020mead} test set for Disgusted and Angry expressions.}
  \label{fig:qualitative_comparison_dis_ang}
\end{figure*}

\begin{figure*}[h]
  \centering
  \includegraphics[width=1\linewidth]{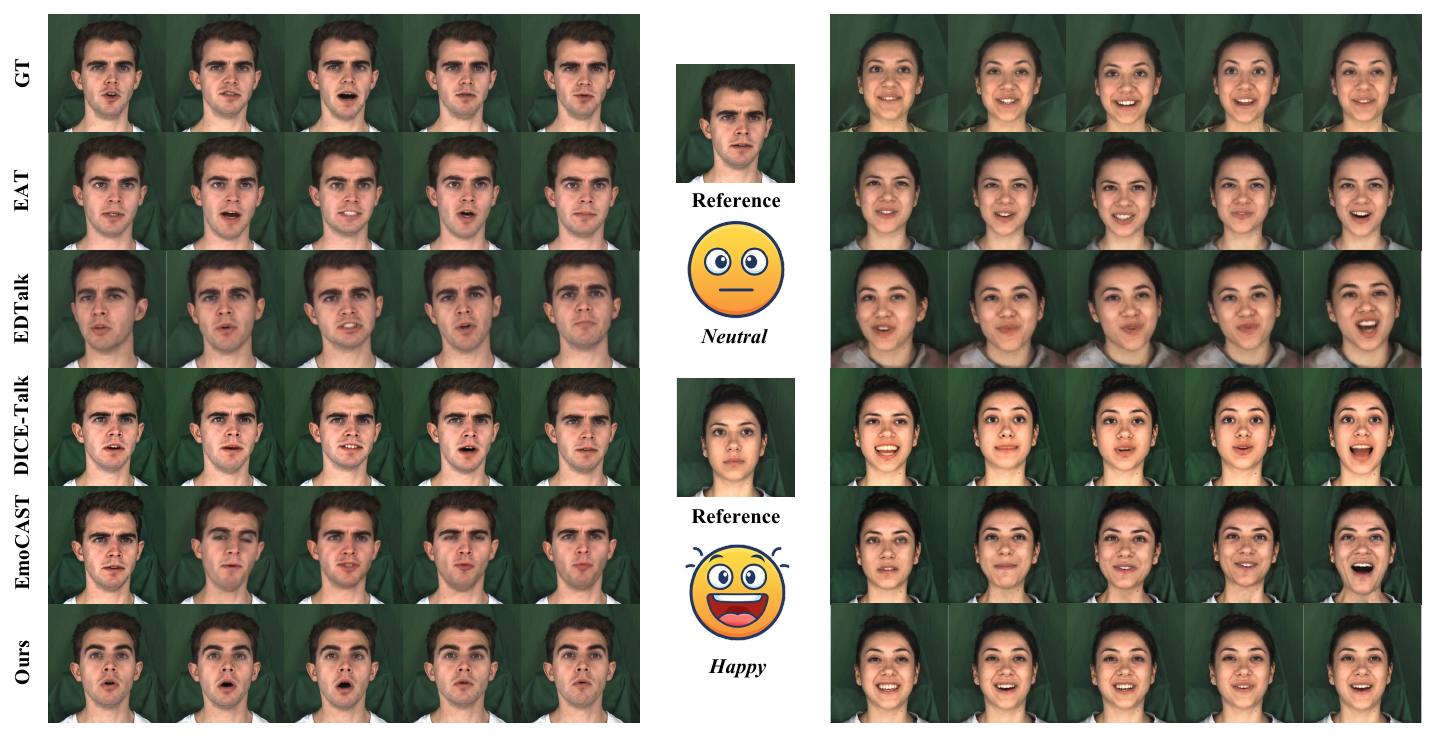}
  \caption{Qualitative comparison on the MEAD~\cite{wang2020mead} test set for Neutral and Happy expressions.}
  \label{fig:qualitative_comparison_neu_hap}
\end{figure*}

\begin{figure*}[h]
  \centering
  \includegraphics[width=1\linewidth]{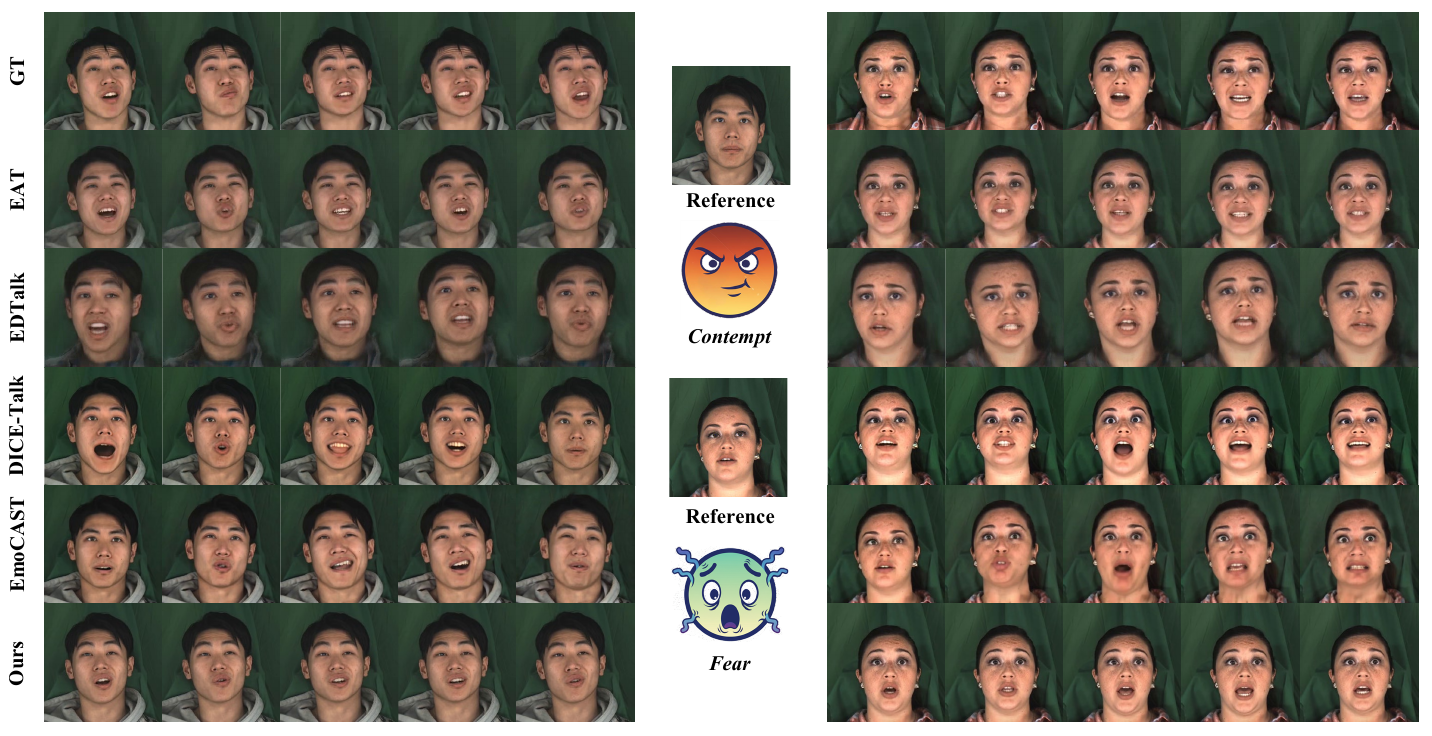}
  \caption{Qualitative comparison on the MEAD~\cite{wang2020mead} test set for Contempt and Fear expressions.}
  \label{fig:qualitative_comparison_con_fea}
\end{figure*}

We provide expanded qualitative comparisons to complement the results presented in the main paper. The additional examples cover a wider range of emotional categories and include both subtle and high-intensity cases. As shown in Fig.\,\ref{fig:qualitative_comparison_dis_ang}, Fig.\,\ref{fig:qualitative_comparison_neu_hap}, and Fig.\,\ref{fig:qualitative_comparison_con_fea}, these examples reveal how different methods behave when facial expressions interact with continuous speech dynamics.

Across all three sets of results, our Xemo-Talker framework consistently produces clearer and more distinguishable emotional cues. The geometry around the mouth corners, eyelids, and nasolabial regions remains coherent throughout the entire sequence, allowing both weak and pronounced expressions to emerge naturally. Compared with other approaches, Xemo-Talker maintains smoother lip trajectories and avoids geometric discontinuities that commonly arise when emotional modulation conflicts with audio-driven articulation. Moreover, identity consistency is preserved even in challenging situations where competing methods show visible drift.

These qualitative observations reinforce the benefits of supervising emotion in the less-principal geometry space. This strategy enables expressive modulation without disturbing articulation patterns, thereby producing emotional variations that remain stable and semantically aligned with the underlying audio.

\section{Sensitivity Analysis}
\label{sec:sensitivity}
\begin{figure*}[t]
  \centering
  \begin{subfigure}[t]{0.32\linewidth}
    \centering
    \includegraphics[width=\linewidth]{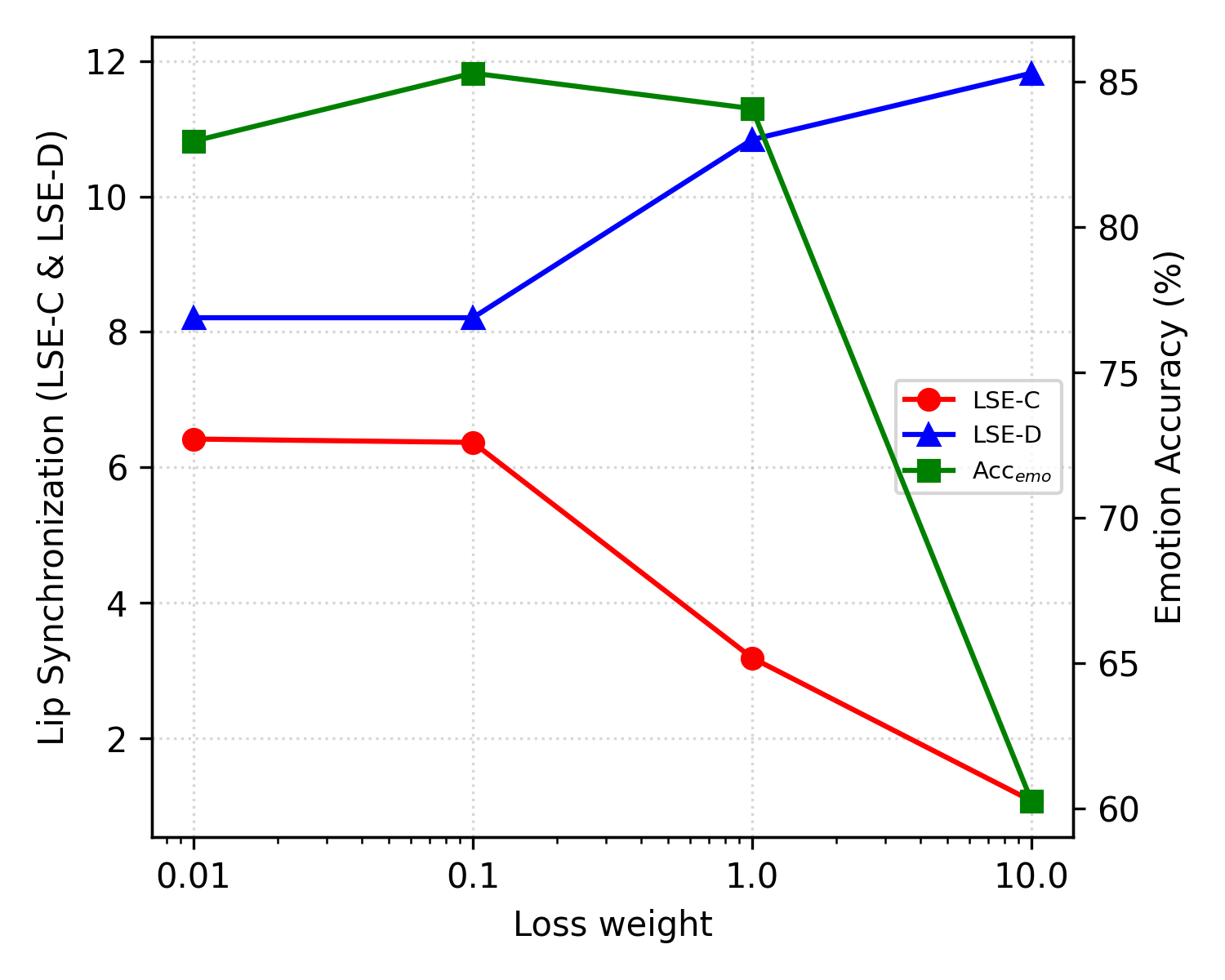}
    \caption{Effect of $\lambda_{\mathrm{lpc}}$ on $L_{\mathrm{lpc}}$.}
    \label{fig:sensitivity_lpc}
  \end{subfigure}
  \hfill
  \begin{subfigure}[t]{0.32\linewidth}
    \centering
    \includegraphics[width=\linewidth]{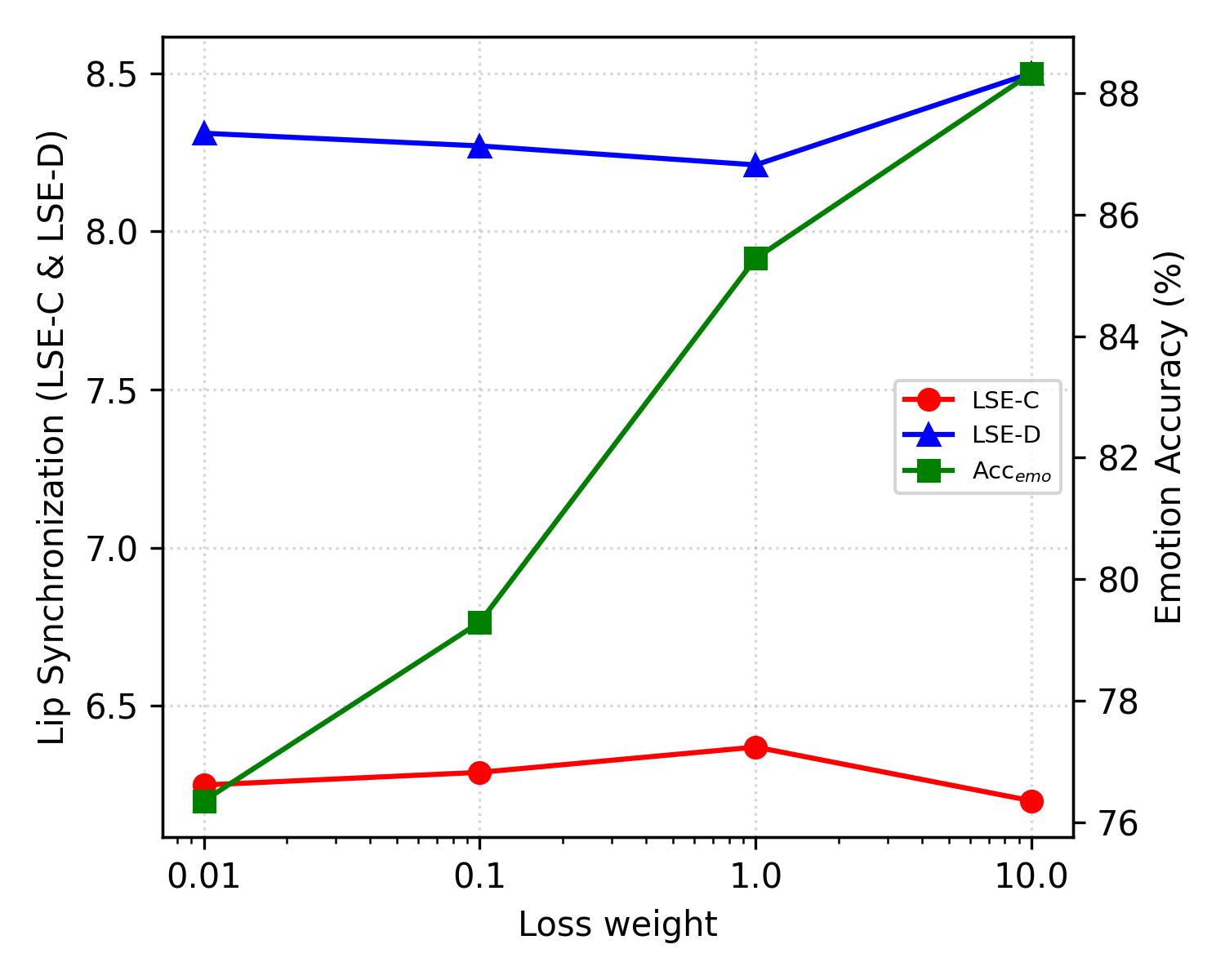}
    \caption{Effect of $\lambda_{\mathrm{cpa}}$ on $L_{\mathrm{cpa}}$.}
    \label{fig:sensitivity_cpa}
  \end{subfigure}
  \hfill
  \begin{subfigure}[t]{0.32\linewidth}
    \centering
    \includegraphics[width=\linewidth]{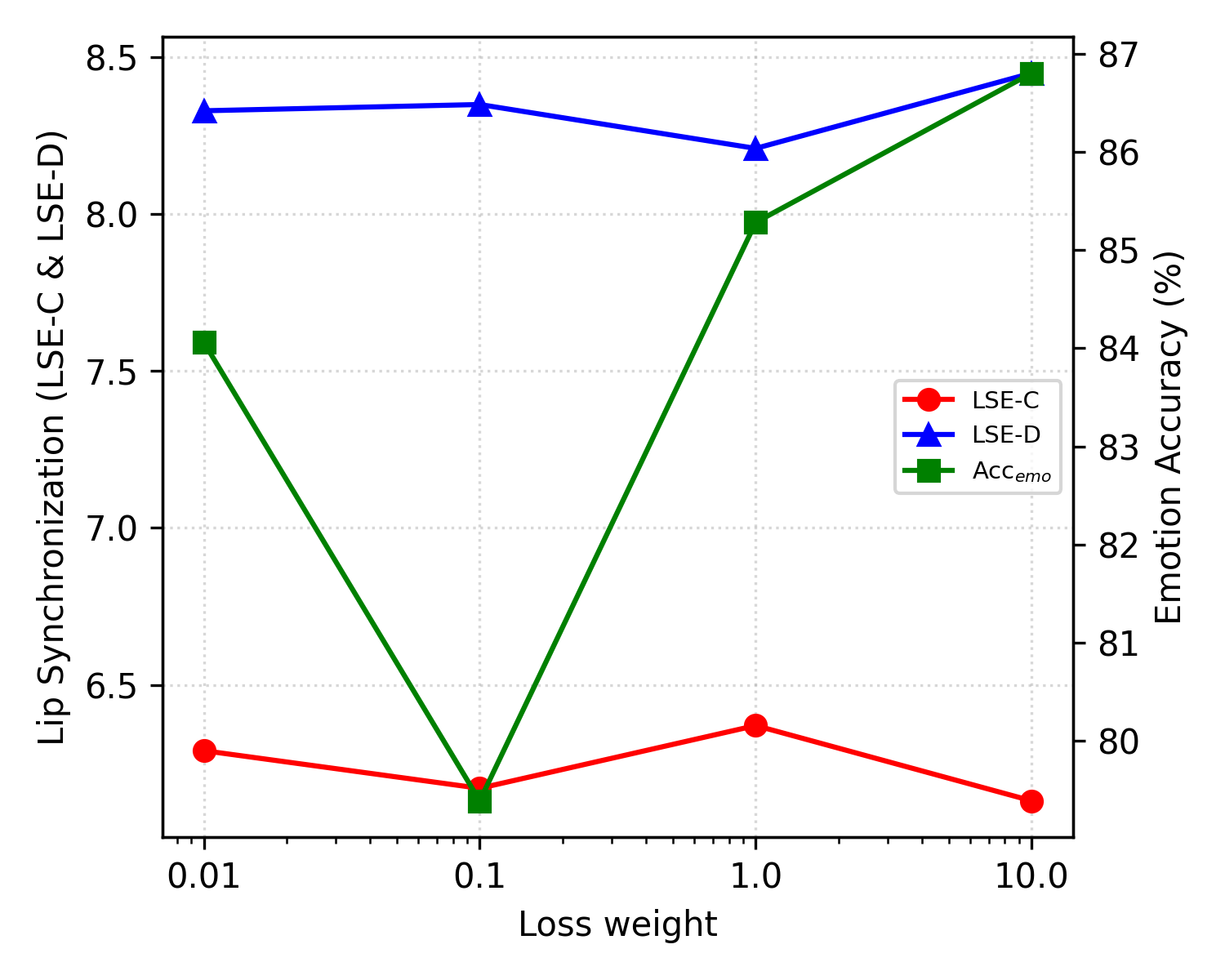}
    \caption{Effect of $\lambda_{\mathrm{cls}}$ on $L_{\mathrm{cls}}$.}
    \label{fig:sensitivity_cls}
  \end{subfigure}
  \caption{Sensitivity study of the three loss weights and their effects on emotion control and articulation quality.}
  \label{fig:sensitivity}
\end{figure*}

To understand how each component of the Tri-Loss influences the final behavior of the model, we conduct a sequential sensitivity study following the procedure illustrated in Fig.\,\ref{fig:sensitivity}. At each step, two loss weights remain fixed while the remaining one is varied, allowing us to isolate and inspect its individual effect on both emotion control and articulation stability.

We begin with the less-principal component loss $L_{\mathrm{lpc}}$ and vary its weight $\lambda_{\mathrm{lpc}}$ while keeping the other two weights unchanged. As shown in Fig.\,\ref{fig:sensitivity_lpc}, smaller values of $\lambda_{\mathrm{lpc}}$ preserve stable lip quality and yield noticeable improvements in emotion accuracy. Larger values introduce excessive emphasis on the less-principal geometry space and can subtly affect lip motion. Based on this trend, we set $\lambda_{\mathrm{lpc}} = 0.1$, which provides the best overall balance.

With $\lambda_{\mathrm{lpc}}$ fixed, we adjust the prototype alignment weight $\lambda_{\mathrm{cpa}}$ and evaluate its influence in Fig.\,\ref{fig:sensitivity_cpa}. This loss encourages emotion features to align with class-specific prototypes. We find that low values weaken class separation and high values reduce within-class variability. A mid-range choice produces the most favorable trade-off, leading to the final selection $\lambda_{\mathrm{cpa}} = 1$.

Finally, keeping both $\lambda_{\mathrm{lpc}}$ and $\lambda_{\mathrm{cpa}}$ fixed, we sweep the classification weight $\lambda_{\mathrm{cls}}$. As illustrated in Fig.\,\ref{fig:sensitivity_cls}, this term stabilizes the global emotion structure and complements prototype alignment. The best performance is obtained at $\lambda_{\mathrm{cls}} = 1$. Together, the three plots in Fig.\,\ref{fig:sensitivity} reveal how each loss contributes to shaping the articulation–emotion balance. They also confirm that the adopted configuration preserves lip synchronization while strengthening emotional expressiveness.

\section{Emotion Ambiguity Analysis}
\label{sec:eaa}
\begin{figure*}[t]
  \centering
  \includegraphics[width=\linewidth]{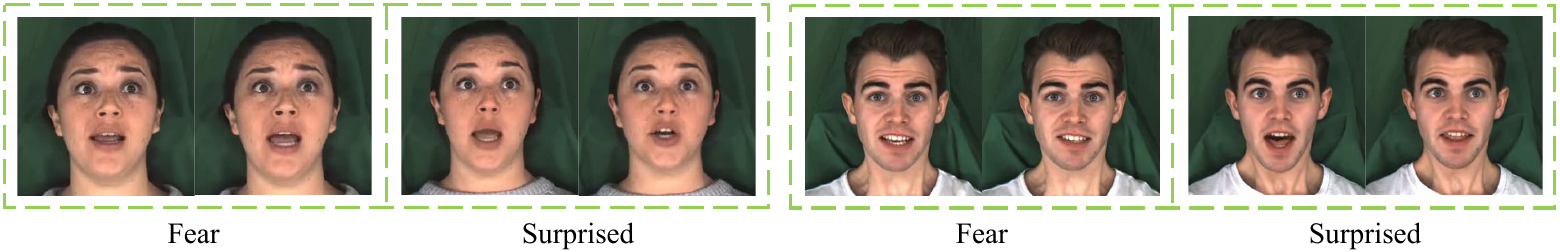}
  \caption{Visual comparison of Fear and Surprised expressions from MEAD~\cite{wang2020mead}.}
  \label{fig:fear_vs_sur}
\end{figure*}

Certain emotion categories in MEAD exhibit highly similar facial configurations. 
As illustrated in Fig.~\ref{fig:fear_vs_sur}, Fear and surprise share characteristic cues such as widened eyes, raised eyebrows, and an open mouth. 
These overlapping facial patterns make the two emotions visually difficult to distinguish even at the appearance level. 
Such intrinsic similarity partially explains the ambiguity observed during emotion evaluation.

\section{User Study Details}
\label{sec:usd}
We conducted a user study based on the MEAD~\cite{wang2020mead} test set. For each of the eight emotions, we selected two subjects, resulting in 16 videos in total. For every sample, we prepared a comparison video that includes the reference image, the target emotion label, and the results from five methods: our model, EAT~\cite{gan2023eat}, DICE-Talk~\cite{tan2025disentangle}, EDTalk~\cite{tan2024edtalk}, and EmoCAST~\cite{jiang2025emocast}. The order of the five results was randomly shuffled for each question to avoid any presentation bias.

A total of 20 participants took part in the study, including 13 men and 7 women. Their backgrounds covered a range of occupations, such as medical staff, software engineers, research assistants, quality inspectors, and students. Each participant was asked to watch the videos and select the best result for three criteria: lip synchronization, emotion accuracy, and overall visual quality. To keep the evaluation fair, all participants viewed the same set of videos, and the ordering of model outputs was independently randomized for every sample. This setup ensures the authenticity and reliability of the collected feedback.

\section{Supplementary Videos}
\label{sec:supplementary_videos}
The supplementary materials include several video results that further demonstrate the effectiveness and robustness of Xemo-Talker across diverse conditions. These videos complement the main paper by showing dynamic behaviors that are difficult to convey through still images.

We first present side-by-side comparisons between our method and competing approaches on the MEAD~\cite{wang2020mead} test set under the standard emotion categories. These videos reveal differences in temporal coherence, emotional transitions, and lip-motion stability. They also highlight how Xemo-Talker responds to both subtle and high-intensity expressions while preserving identity over extended sequences.

We then show additional results on out-of-domain data to assess generalization. The sources span varied identities, camera setups, lighting conditions, and recording environments. These videos demonstrate that the geometry-based design and the less principal emotion supervision transfer reliably beyond MEAD~\cite{wang2020mead}, maintaining stable articulation and clear emotional cues under distribution shifts.

Next, we include emotion interpolation results. Given a fixed audio clip, we gradually change the target emotion from angry to happy. The sequence exhibits smooth, continuous transitions while maintaining accurate lip motion, illustrating the controllability enabled by the geometry emotion branch and the Tri-Loss design.

We also provide a PCA-based visualisation comparing emotion supervision applied to different PCA regions. Each clip shows three results: supervision on the full space, on a head slice of principal components, and on the tail slice covering the low-variance region. The video shows that full-space and head-region supervision can distort mouth motion or overdrive expression, whereas tail-region supervision mainly adjusts subtle details and better preserves articulation.

Finally, we include a loss ablation video showing the effects of removing individual terms. By comparing sequences without classification, prototype alignment, or low-variance contrastive objectives, the video highlights their contributions to expression clarity and the balance between emotional control and articulation quality.

\end{document}